\documentclass{article}
\usepackage{arxiv}
\usepackage[utf8]{inputenc} 
\usepackage[T1]{fontenc}    
\usepackage{url}            
\usepackage{cite}
\usepackage{amsmath,amssymb,amsfonts}
\usepackage{algorithmic}
\usepackage{graphicx}
\usepackage{textcomp}
\usepackage{times}
\usepackage{epsfig}
\usepackage{graphicx}
\usepackage{amsmath}
\usepackage{amssymb}
\usepackage{caption}
\usepackage{sidecap}
\usepackage{tikz}
\usetikzlibrary{intersections}
\usetikzlibrary{shapes.geometric}
\usepackage{mathtools}
\usepackage{pgfplots, pgfplotstable}
\usepgfplotslibrary{statistics}
\usepackage{textcomp}
\usepackage{soul}
\usepackage{multirow}
\usetikzlibrary{spy}
\usepackage{hyperref}
\usepackage{scalerel}
\usepackage{caption}
\usepackage{multirow}
\usepackage{microtype}
\usepackage{graphicx}
\usepackage{subfigure}
\usepackage{tikz}
\usetikzlibrary{spy}
\usetikzlibrary{calc,arrows.meta,positioning}
\usepackage{pgfplots}
\pgfplotsset{compat=1.18}
\usetikzlibrary{shapes.geometric}
\usepackage{pgfplotstable}
\usepackage{xcolor}
\usepackage{booktabs}
\def\BibTeX{{\rm B\kern-.05em{\sc i\kern-.025em b}\kern-.08em
    T\kern-.1667em\lower.7ex\hbox{E}\kern-.125emX}}

\title{ProARD: Progressive Adversarial Robustness Distillation: Provide Wide Range of
Robust Students}

\date{} 				

\author{
   Seyedhamidreza Mousavi\\
	Mälardalen University\\
	\texttt{seyedhamidreza.mousavi@mdu.se} \\
	\And
	\hspace{1mm}Seyedali Mousavi\\
	Mälardalen University\\
	\texttt{seyedali.mousavi@mdu.se} \\
	\AND
	\hspace{1mm}Masoud Daneshtalab \\
        Mälardalen University\\
	\texttt{masoud.daneshtalab@mdu.se}
}

\date{}

\renewcommand{\shorttitle}{ProARD: Progressive Adversarial Robustness Distillation: Provide Wide Range of Robust Students}
\begin{document}
\maketitle
\begin{abstract}
Adversarial Robustness Distillation (ARD) has emerged as an effective method to enhance the robustness of lightweight deep neural networks against adversarial attacks. 
Current ARD approaches have leveraged a large robust teacher network to train one robust lightweight student. 
However, due to the diverse range of edge devices and resource constraints, current approaches require training a new student network from scratch to meet specific constraints, leading to substantial computational costs and increased CO\textsubscript{2} emissions.

This paper proposes Progressive Adversarial Robustness Distillation (ProARD)\footnote{https://github.com/hamidmousavi0/ProARD}, enabling the efficient one-time training of a dynamic network that supports a diverse range of accurate and robust student networks without requiring retraining. 
We first make a dynamic deep neural network based on dynamic layers by encompassing variations in width, depth, and expansion in each design stage to support a wide range of architectures ($>10^{19}$).
Then, we consider the student network with the largest size as the dynamic teacher network. 
ProARD trains this dynamic network using a weight-sharing mechanism to jointly optimize the dynamic teacher network and its internal student networks.
However, due to the high computational cost of calculating exact gradients for all the students within the dynamic network, a sampling mechanism is required to select a subset of students. 
We show that random student sampling in each iteration fails to produce accurate and robust students.
ProARD employs a progressive sampling strategy that gradually reduces the size of student networks in three steps during training while applying robustness distillation between the dynamic teacher network and the selected students.
Finally, we leverage a multi-objective evolutionary algorithm based on a proposed accuracy-robustness predictor to identify optimal architectures that balance accuracy, robustness, and efficiency.

Through the experiments, we show that ProARD reduces the computational cost by $60 \times$ and improves accuracy and robustness by $13\%$ and $14\%$, respectively, compared to random sampling. 
We also demonstrate that our accuracy-robustness predictor can estimate the accuracy and robustness of test student networks with root mean squared errors of $0.0073$ and $0.0072$, respectively. 
%
%
\end{abstract}
\section{Introduction}
\label{sec:intro}
Deep Neural Networks (DNNs) are highly effective in solving complex tasks, including image classification~\cite{krizhevsky2012imagenet}, object detection~\cite{zou2023object}, and image segmentation~\cite{zou2024segment}.
Recently, DNNs have been widely used in security-critical applications, such as self-driving cars~\cite{eykholt2018robust}, face recognition~\cite{sharif2016accessorize}, and medical diagnosis~\cite{ma2021understanding}, where robustness to input perturbations is the primary concern. 
Despite their strengths, DNNs are susceptible to subtle and imperceptible changes in input data, referred to as adversarial attacks~\cite{szegedy2013intriguing}. These vulnerabilities pose significant challenges for security-critical applications. 

Different strategies have been suggested to counter adversarial attacks and enhance the robustness of DNNs.
Among them, Adversarial Training (AT) is currently the most effective method for robustification~\cite{zhang2019theoretically}. 
AT methods aim to train an adversarially robust DNN such that its predictions are locally invariant to a small neighborhood of inputs.
However, AT requires a well-designed, high-capacity network architecture, and its effectiveness is limited in lightweight networks due to their limited capacity~\cite{ou2024towards}.
To address this issue, the Adversarial Robustness Distillation (ARD) technique has been designed to transfer robustness from a high-capacity robust teacher network to a lightweight student network~\cite{dong2024robust}. 
Current ARD methods are limited to training a single student network and primarily focus on finding the best training loss function based on the alignment between teacher and student networks\cite{dong2025adversarially}.
Consequently, addressing the heterogeneity of edge devices and resource constraints requires retraining various network architectures tailored to each deployment scenario. 
Therefore, as the number of deployment scenarios ($N$) increases, the total cost of designing a lightweight network grows linearly (i.e. $O(N)$).
This approach poses significant challenges for practical implementation across varying hardware configurations, while also incurring substantial computational costs and contributing to increased CO\textsubscript{2} emissions.
Given this limitation, we pose the following question: 
\textit {How can we take advantage of adversarial robustness distillation to provide a wide range of robust students efficiently without retraining them?} 

This paper introduces a Progressive Adversarial Robustness Distillation (ProARD) training strategy to train a wide range of student network architectures suitable for different deployment scenarios.
ProARD makes a dynamic deep neural network based on dynamic layers, including variations in width (kernel size), depth, and expansion in each design stage to support a wide range of architectures.
To train this dynamic network, we consider the largest configuration of the dynamic network as the dynamic teacher and jointly train the dynamic teacher and students in a weight-sharing mechanism.
However, due to the huge number of students in the dynamic network, it is computationally prohibitive to compute the exact gradient based on all students (more than $10^{19}$ students). 
We address this by sampling a subset of students by progressively reducing their size in each training iteration, and we show that this method works better than random sampling. 
After training the dynamic network by ProARD, we utilize a multi-objective evolutionary search algorithm alongside an accuracy-robustness predictor to identify the optimal student effectively.
Our contributions can be summarized as follows: 
\begin{itemize}
    \item We present a novel method called Progressive Adversarial Robustness Distillation (ProARD)\footnote{https://github.com/hamidmousavi0/ProARD}, designed to train a dynamic network including a wide range of students by progressive sampling.
    \item We propose a joint accuracy-robustness predictor to efficiently estimate the robustness and accuracy, leveraging it in the multi-objective evolutionary algorithm to quickly find the best architecture based on the resource constraints without retraining. 
    \item We extensively evaluate the distribution of the robustness and accuracy of students and the effectiveness of our accuracy-robustness predictor and multi-objective search mechanism on the CIFAR-10 and CIFAR-100 datasets with the dynamic networks based on ResNet and MobileNet architectures. 
\end{itemize}
\section{Related Work}
\label{sec:related_work}
\subsection{Adversarial Attacks \& Training}
\label{sec:related_work:Adversarial Attacks and Defense}
Adversarial attacks create subtle alterations in input data to intentionally mislead DNNs into making incorrect predictions. Common techniques for generating these attacks include the Fast Gradient Sign Method (FGSM) \cite{goodfellow2014explaining}, Projected Gradient Descent (PGD) \cite{madry2017towards}, and the Carlini \& Wagner (CW) attack~\cite{carlini2017towards}, among others~\cite{costa2024deep}. Furthermore, AutoAttack (AA)~\cite{croce2020reliable}, a set of four techniques combining white-box and black-box methods, has shown remarkable effectiveness in generating adversarial attacks.
Adversarial training is the current state-of-the-art defense method against adversarial attacks, aiming to create robust DNNs.
The earliest adversarial training approach leveraged both clean and adversarial images to train a robust DNN~\cite{goodfellow2014explaining}.
%
%
Adversarial training can be reformulated as a min-max optimization problem, where the network is trained exclusively on adversarial examples~\cite{madry2017towards}.
TRADES~\cite{zhang2019theoretically} regularizes the loss function for clean data by incorporating a robust loss term and making a trade-off between them.
TRADES variants have also been proposed to consider regularization terms and reduce the distance between the distribution of natural data and their adversarial counterparts ~\cite{cui2021learnable,wang2020improving}.
Although adversarial training methods work well for high-capacity network architectures, their effectiveness is limited in lightweight networks due to their limited capacity~\cite{ou2024towards}.
%
%
%
\subsection{Adversarial Robustness Distillation}
%
Adversarial Robustness distillation (ARD)~\cite{goldblum2020adversarially, zi2021revisiting,zhu2021reliable, jung2024peeraid, yin2024adversarial} is a knowledge distillation technique designed to transfer robustness from a large, robust network to a smaller, more efficient network.
ARD~\cite{goldblum2020adversarially} has shown that robustness distillation can produce a student with greater robustness than training from scratch. It encourages the student network to mimic the teacher logits within an $\epsilon$-ball around the data points.
IAD~\cite{zhu2021reliable} focuses on the reliability of the teacher and the student network trust in the teacher network based on the performance of the student on average and natural data. 
RSLAD~\cite{zi2021revisiting} introduces the concept of robust soft labels (RSL) generated by the robust teacher, providing an effective and robust representation of the student network. 
PeerAiD~\cite{jung2024peeraid} is an adversarial distillation approach that simultaneously trains the peer network and the student network, enabling the peer network to specialize in defending the student network.
MTARD~\cite{zhao2022enhanced} utilizes multiple teachers to guide smaller networks in an adversarial setting.
AdaAD~\cite{huang2023boosting} introduces an adaptive knowledge distillation method that also considers the teacher in the optimization process.
SmaraAD~\cite{yin2024adversarial}  aligns the attribution regions of the student with those of the teacher network, facilitating a closer correspondence between the outputs of the teacher and student networks. 
Although adversarial robustness distillation methods provide a robust lightweight student network, these methods provide a single robust student network at each run and require the training process to be repeated for each scenario with different resource constraints.
ProARD introduces a progressive robustness distillation approach to develop a dynamic robust network that encompasses a diverse set of robust student networks, adaptable to various deployment scenarios. 
ProARD works independently of specific robustness distillation methods; in this paper, we adopt RSLAD~\cite{zi2021revisiting} as the baseline.

\section{Research Motivation}
\begin{figure}[t]
\centerline{\includegraphics[width=0.5\textwidth]{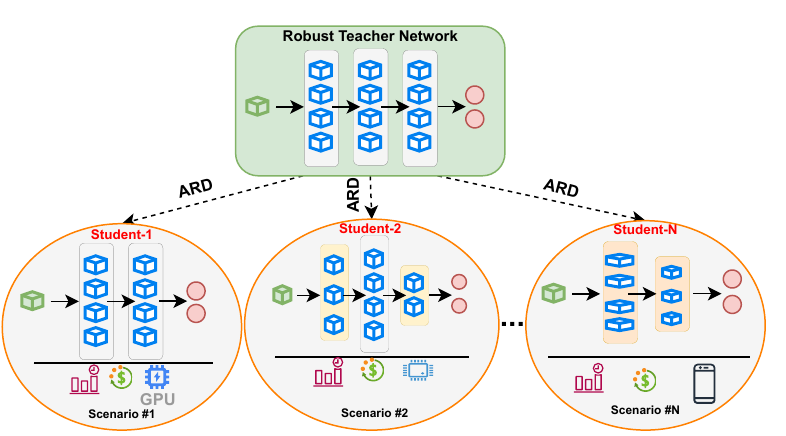}}
\caption{Adversarial Robustness distillation(ARD) for different deployment scenarios (GPU, CPU, Mobile, etc) }
\label{fig1:robustdistillation}
\end{figure}
Adversarial robustness distillation methods utilize a robust teacher network to train a robust student.
However, the training process must be repeated for a variety of edge devices and deployment scenarios, which is both time-intensive and resource-demanding.
As shown in Fig.~\ref{fig1:robustdistillation}, robustness distillation for different deployment scenarios requires retraining for each hardware device and set of resource constraints, resulting in significant computational costs and substantial CO\textsubscript{2} emissions.

\begin{figure}[htbp]
    \centering
    \resizebox{\columnwidth}{!}{
         \begin{tikzpicture}
     \begin{axis}[
            width=0.7\columnwidth,
            height=0.5\columnwidth,
             legend style={at={(0.141,1)},
              anchor=north,legend columns=1,font=\tiny},
            font=\tiny,
            scaled x ticks = true,
            scaled y ticks = true,
            ymin=35, ymax=55,
            xmin =65, xmax=87,
            grid=major, 
            grid style={dashed,gray!40}, 
            ylabel near ticks,
            xlabel near ticks,
            xlabel=  \textbf{Accuracy(\%)}, 
            ylabel=  \textbf{Robustness(\%)},
            ]

            \addplot [red,only marks,mark size=1pt] table[x=acc, y=rob, col sep=comma] {wps_csv_resnet50.csv};
            \addplot [blue,only marks,mark size=1pt] table[x=acc, y=rob, col sep=comma] {csv.csv};
            \addplot [fill, mark=pentagon*, only marks, mark size=2.5pt, green, thick,draw=black]  coordinates {(84.7,52.6)};
            \legend{Random,\textbf{ProARD},ResNet-50}
        \end{axis}
        \end{tikzpicture}}
    \caption{The accuracy-robustness distribution of students when trained with random sampling (red) and our ProARD (blue). For evaluation, we use the PGD attack with  $\epsilon=0.031$ and step-size $=0.0078$. The green point shows the accuracy and robustness of the ResNet-50 on the CIFAR-10 dataset.}
    \label{fig:dist-WP}
\end{figure}
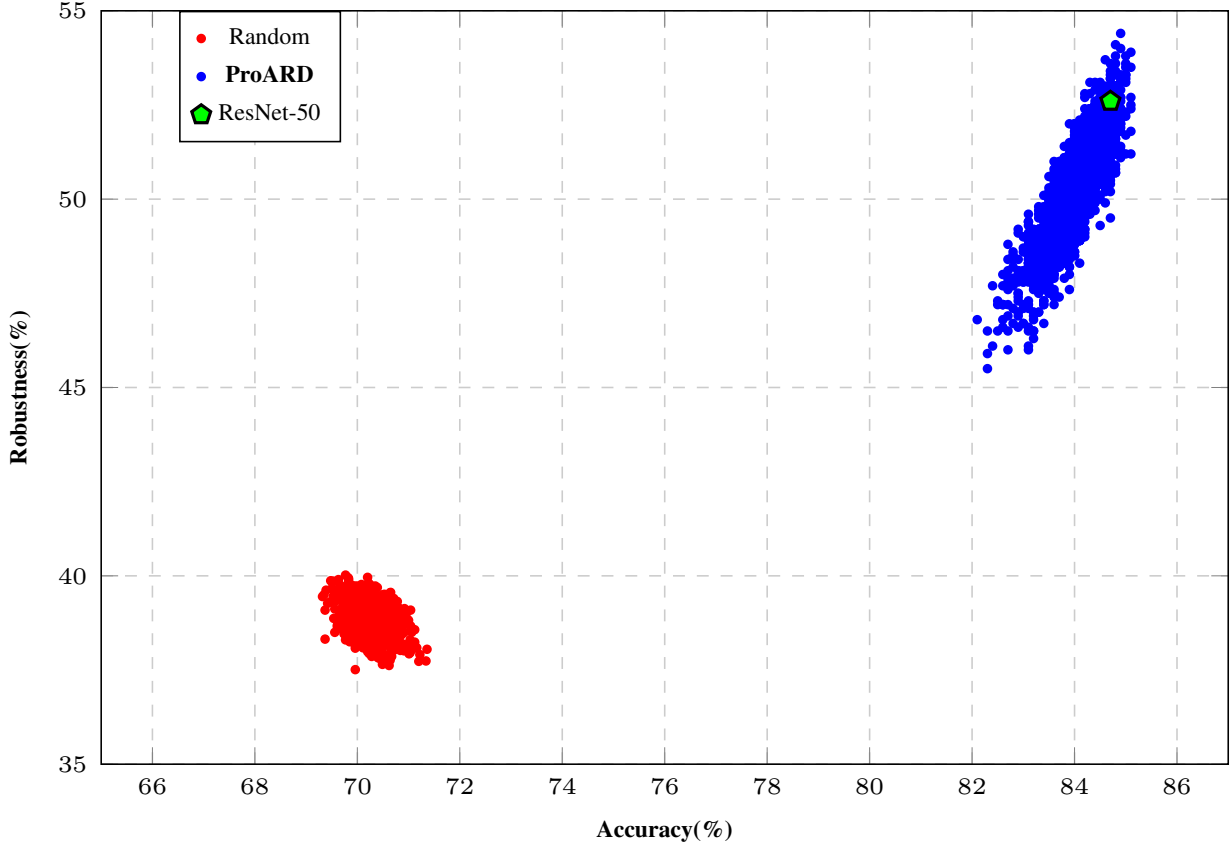

To address this limitation, the primary motivation of ProARD is to train a robust dynamic network from which multiple student networks can be extracted without requiring retraining.
To achieve this, we design a dynamic robust network with dynamic layers capable of varying widths (kernel sizes), depths, and expansion, enabling flexibility in its structure.
The dynamic networks are built using bottleneck residual blocks (Dynamic ResNet) and inverted bottleneck blocks (Dynamic MobileNet), incorporating diverse configuration parameters at each design stage.
For training, we first adversarially train the largest student network within the dynamic network using the TRADE~\cite{zhang2019theoretically} method.
This largest network then serves as a dynamic teacher, guiding the training of randomly sampled student networks in each iteration.
The student networks share their weights with the dynamic teacher network to ensure consistency.
Given the enormous number of possible student networks within the dynamic network (exceeding $10^{19}$ networks), it is computationally infeasible to calculate exact gradients for all networks and share their weights.
To overcome this, we randomly sample student networks during each training iteration.
The red points in Fig.~\ref{fig:dist-WP} depict the accuracy-robustness distribution of $2,000$ student networks trained via random sampling during each iteration within the Dynamic ResNet network on the CIFAR-10 dataset.
The green point represents the robustness and accuracy of the ResNet-50 base network, which is adversarially trained using the TRADES\cite{zhang2019theoretically} algorithm on the CIFAR-10 dataset.
Most student networks extracted from the dynamic network trained using the random sampling strategy exhibit significantly lower accuracy and robustness compared to ResNet-50.
The most accurate student network has $13.34\%$ and $12.58\%$ lower accuracy compared to ResNet-50. 
This indicates that the random sampling and weight-sharing strategy for training the dynamic network fails to produce robust and accurate student networks.
To address this shortcoming, a new training strategy is necessary to enhance the robustness and accuracy of student networks.
ProARD introduces a novel progressive robustness distillation approach designed to overcome these limitations and improve the performance of student networks. 
The blue points in Fig.~\ref{fig:dist-WP} show the distribution of $2,000$ student networks trained via our proposed method (ProARD) that can provide accurate and robust students.

\section{Method}
\begin{figure*}[htbp]
\centerline{\includegraphics[width=\textwidth]{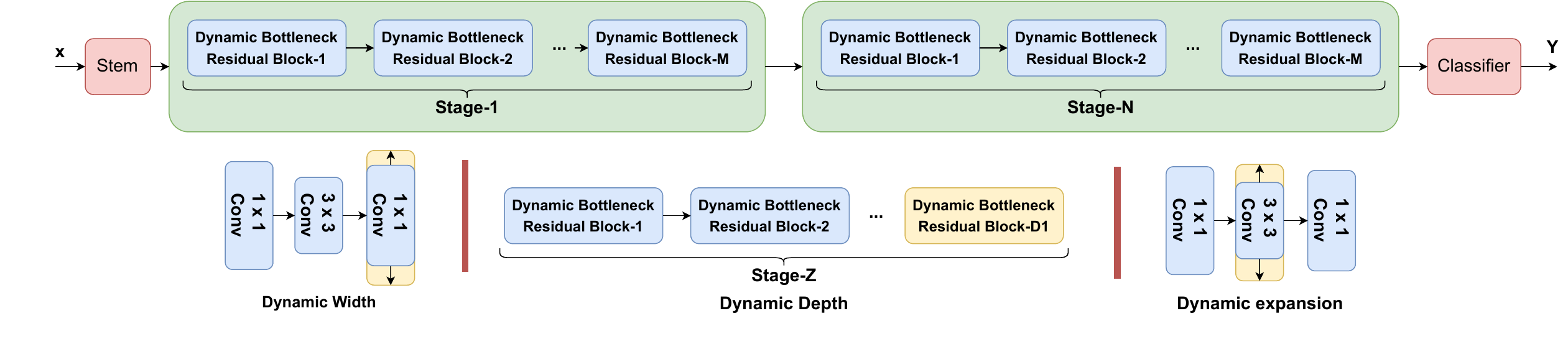}}
\caption{The architecture of Dynamic ResNet with dynamic bottleneck residual blocks. Each dynamic bottleneck residual block has dynamic width and expansion and each stage has a dynamic depth. }
\label{fig1:DynamicResNet50}
\end{figure*}
\subsection{Preliminaries}
Adversarial training can be formulated as the following min-max optimization problem:
\begin{equation}\label{Adv-training}
    \begin{split}
    & \underbrace{\min_{\textbf{w}} \mathbb{E}_{(x,y)\sim P} \mathcal{L}_{\min}(f_{\textbf{w}}(x+\delta),y)}_{\textbf{Outer minimization}}\\
    & \text{s.t.}\;\;\ \delta =  \underbrace{\max_{||\delta||_p\le\epsilon}\mathbb{E}_{(x,y)\sim P} \mathcal{L}_{\max}(f_{\textbf{w}}(x+\delta),y)}_{\textbf{Inner maximization}}
    \end{split}
\end{equation}
Where $P$ is the data distribution, $(x\in \mathcal{X},y\in \mathcal{Y})$ is a pair of input features and labels, $f_{\textbf{w}}(.)$ is the DNN  with parameters $\textbf{w}$, and $\delta$ is the perturbation within the bounded $l_p$ distance ($\epsilon$) . $\mathcal{L}_{\min}$ and $\mathcal{L}_{\max}$ are the loss functions for the inner and outer optimization problems, respectively.
%
%
The Cross-Entropy (CE) loss function is specified as $\mathcal{L}_{\min}$, while $\mathcal{L}_{\max}$ may vary depending on the chosen method.
For example, SAT~\cite{madry2017towards} and TRADE~\cite{zhang2019theoretically} methods employ cross-entropy loss and KL divergence as $\mathcal{L}_{\max}$.

Adversarial robustness distillation (ARD) has been introduced to take advantage of the performance of a robust teacher network. 
It first trained a large capacity teacher network $T(.)$ with adversarial training following Eq.~\ref{Adv-training}, then the teacher used to provide soft labels with natural data ($x_i$) and adversarial example ($x_i+\delta$) of the student network $S(.)$. 
By using different loss functions for $\mathcal{L}_{\min}$ and $\mathcal{L}_{\max}$, we can formulate different robustness distillation methods.
However, they can train only one student network, and the distillation process must be repeated for each deployment scenario. 
Our method is independent of the robustness distillation methods and we used the RSLAD~\cite{zi2021revisiting} adversarial distillation which considers the following loss functions for $\mathcal{L}_{\min}$ and $\mathcal{L}_{\max}$: 
\begin{equation}\label{RSLAD_L}
\begin{split}
    \mathcal{L}_{\min} = (1-\alpha)& \cdot \boldsymbol{KL}(S(x),T(x)) + \alpha \cdot \boldsymbol{KL}(S(x+\delta),T(x))   \\
    & \mathcal{L}_{\max} = \boldsymbol{KL}(S(x+\delta),T(x))    
\end{split}   
\end{equation}
Where $\boldsymbol{KL}$ is the KL-divergence. $S(x)$ and $T(x)$ indicate the student and teacher with parameters $\boldsymbol{w_s}$ and $\boldsymbol{w_t}$ respectively.

\subsection{Problem Formulation} 
DNNs are organized into multiple design stages, with each stage consisting of a specific arrangement of layers.
A dynamic network architecture can be constructed by varying configuration parameters such as width (kernel size), depth, and expansion across these stages and layers. 
Assuming the largest dynamic network architecture, configured with the maximum values for these parameters, serves as the dynamic teacher with trainable parameters $\boldsymbol{w_{t}}$. 
The dynamic network encompasses a diverse set of student architectural configurations, denoted as $\mathcal{S}$, from which various student networks $s$ can be selected. 
The primary objective is to train a robust dynamic teacher network and utilize a robustness distillation method to ensure robustness across all students.  
To achieve a wide range of robust student networks, we aim to solve the following optimization problem:
\begin{equation}\label{adv_dyn}
    \begin{split}
        &\min_{\boldsymbol{w_t}} \mathbb{E}_{s\in \mathcal{S}}\mathbb{E}_{(x,y)\sim P}\mathcal{L}_{\min}(s^{\boldsymbol{w_t^s}}(x+\delta),y)\\
        & \text{s.t.}\;\;\ \delta =  \max_{||\delta||_p\le\epsilon}\mathbb{E}_{(x,y)\sim P} \mathcal{L}_{\max}(s^{\boldsymbol{w_t^s}}(x+\delta),y))
    \end{split}
\end{equation}
Where $s^{\boldsymbol{w_t^s}}$ refers to trainable parameters of the student network chosen according to the $s$ configuration and a specific selection scheme. 
The primary goal of the training process is to adversarially optimize the dynamic network so that every student architecture achieves a similar level of accuracy and robustness.
To solve the optimization problem (~\ref{adv_dyn}), two approaches can be considered.
The first approach involves selecting and training all the student networks within the dynamic network from scratch.
However, with over $10^{19}$ possible student networks, this solution is computationally infeasible.
Moreover, as demonstrated in the motivation section, randomly selecting student networks for training fails to achieve satisfactory accuracy and robustness.
To address these challenges, a novel method is required to train both the dynamic network and its student networks effectively.
The proposed solution involves constructing a dynamic network and progressively selecting student networks from it during each training iteration, as described in the following sections.
\begin{figure*}[htbp]
\centerline{\includegraphics[width=0.9\textwidth]{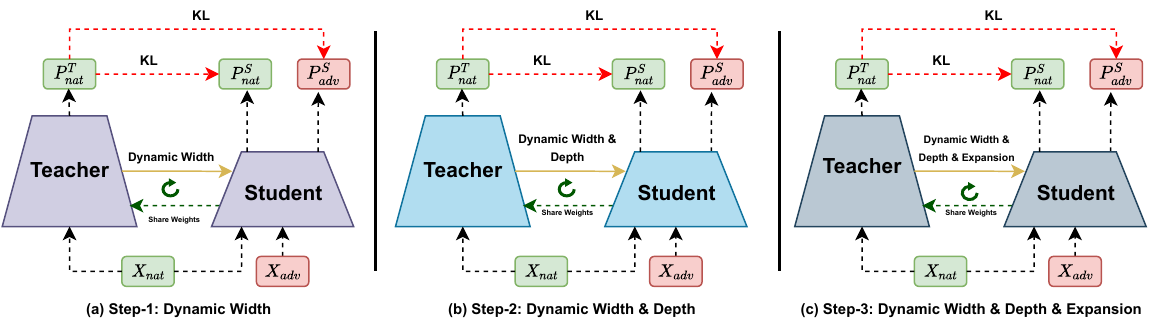}}
\caption{ProARD: Three steps Progressive Adversarial Robustness Distillation Framework. (a) Step-1: train dynamic width (b) Step-2: train dynamic width and depth, and (c) Step-3: train dynamic width, depth, and expansion.}
\label{fig1:ProARD}
\end{figure*}
\subsection{Dynamic Network Architecture}
To support a diverse range of student network architectures, we made a dynamic network based on dynamic layers which cover different networks based on three dimensions (i.e., depth, width (kernel size), and expansion) in a deep neural network.
We construct a Dynamic ResNet by incorporating variations in width, depth, and expansion in bottleneck residual blocks, while we make Dynamic MobileNet by utilizing kernel size, depth, and expansion in inverted bottleneck blocks.
Following the common practice of DNN networks, we divide each DNN into a sequence of different stages with multiple layers that gradually reduce the size of the feature map and increase the number of channels. 
To make a dynamic network architecture, we allow each stage to have an arbitrary number of depths (dynamic depth). 
We allow each layer to have an arbitrary number of channels, arbitrary width (for Dynamic ResNet) and kernel size (for Dynamic MobileNet) (dynamic expansion, dynamic width, and dynamic kernels) in each stage.
Fig~\ref{fig1:DynamicResNet50} indicates the architecture of our Dynamic ResNet based on the dynamic bottleneck residual blocks.
We can consider different widths and expansions by altering the number of output channels in the last conv and middle conv layers in the bottleneck residual block and using different depths by reducing the number of blocks in each stage.  
With this design for a DNN with 5 stages, if we select the depth of each stage from $\{2,3,4\}$ and use $3$ different values for the width and expansion of each layer in the stages, then we have roughly $((3 \times 3)^2 + (3\times3)^3 + (3\times3)^4)^5 \approx 2 \times 10^{19}$ student network architectures in our dynamic network.  
The architecture of Dynamic MobileNet is similar to that in Fig~\ref{fig1:DynamicResNet50}, using inverted bottleneck blocks. The only modification is replacing the dynamic width with a dynamic kernel size in this network.
\subsection{Progressive Adversarial Robustness Distillation}
To efficiently train student networks within the dynamic network, we propose Progressive Adversarial Robustness Distillation (ProARD), which enables the joint training of the dynamic teacher network—the largest student network within the dynamic architecture—and a wide range of student networks.
In ProARD, we start by selecting the maximum values for the configuration parameters of the dynamic network to construct the largest network. 
%
%
We adversarially train the largest network and utilize it as the dynamic teacher network for distillation.
To overcome the sampling challenges discussed in the motivation section, we adopt a progressive adversarial robustness distillation approach that progressively samples student network architectures from large to small in three steps.
In the first step, we fix the depth and expansion, extracting different student networks by varying the width (kernel size) of the bottleneck residual block (inverted bottleneck block) in each iteration. 
Robustness distillation based on the RSLAD (Eq~\ref{RSLAD_L}) is applied between the dynamic teacher and students.
After training, the student parameters are shared with the dynamic teacher network.   
In the second step, we fix the expansion while varying the width (kernel size) and depth to train student networks using robustness distillation, then share the weights with the dynamic teacher network.
In the third step, we extract student networks by varying all three parameters, train them, and apply the weight-sharing.
Fig~\ref{fig1:ProARD} illustrates the progressive robustness distillation mechanism used to build a dynamic network that supports a wide range of robust and lightweight student networks. 
\begin{figure*}[htbp]
\centerline{\includegraphics[width=0.9\textwidth]{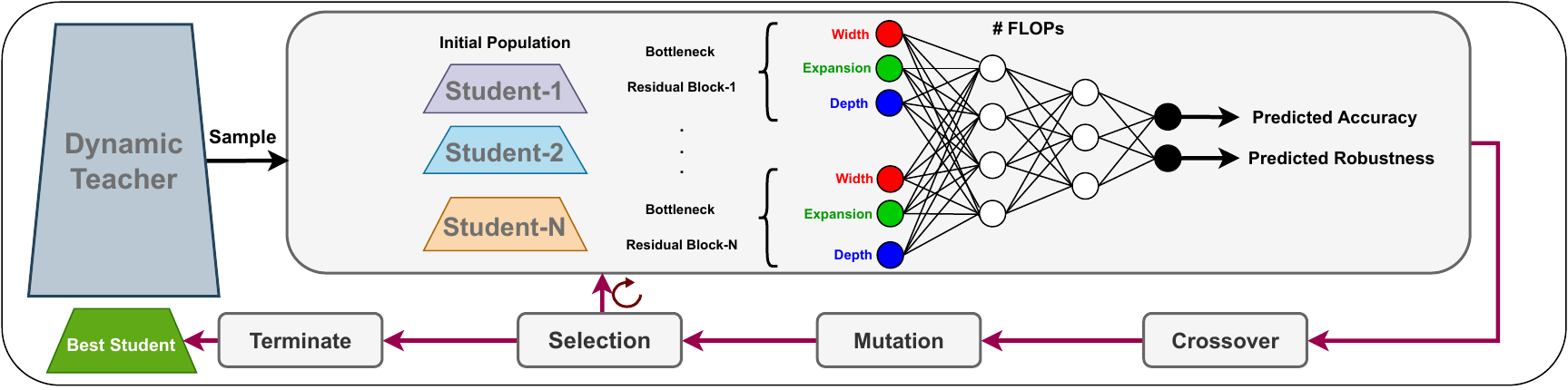}}
\caption{Multi-objectice Search engine consists of our proposed accuracy-robustness predictor}
\label{fig1:NSGAII}
\end{figure*}
\subsection{Multi-objective Search}
After training the dynamic network using ProARD, we obtain a diverse set of robust student networks within the dynamic teacher network.
The next step is to identify the best student network based on the given resource constraints and deployment scenario.
We employ a multi-objective search mechanism that takes robustness, accuracy, and efficiency (i.e., number of FLOPs) into account to identify the optimal student network.
However, computing robustness and accuracy during the search process is time-consuming, making it impractical to evaluate robustness for a wide range of students.
To address this challenge, we propose an accuracy-robustness predictor network that provides quick feedback for the search process.
Specifically, we sample $2K$ student network architectures from the dynamic network and measure their robustness and accuracy by directly evaluating them.
This dataset ([student architecture, robustness, accuracy]) is then used to train a predictor that can estimate the robustness and accuracy of any given student network.
For the specification of the student architectures, we leverage the width (kernel size), depth, and expansion values in each stage of design and layers. 
With the predictor in place, the multi-objective search process can efficiently find the best student architecture for a given deployment scenario, considering FLOPs, robustness, and accuracy.
By training the predictor just once, we reduce the search cost, and it remains constant regardless of the deployment scenario.
Fig~\ref{fig1:NSGAII} illustrates the architecture of our accuracy-robustness predictor and the multi-objective search to identify the best student network. 

\section{Exprimental Results} \label{ER}
In this section, we compare random sampling-based robustness distillation training with our proposed ProARD approach for addressing the optimization problem outlined in Eq.~\ref{adv_dyn}.
We then demonstrate the effectiveness of our accuracy-robustness predictor within the multi-objective evolutionary algorithm, showing that our method can efficiently identify the optimal student.
Additionally, we assess the adversarial robustness and accuracy of the optimal student architecture under various white-box attacks.
We use TRADES~\cite{zhang2019theoretically} and RSLAD~\cite{zi2021revisiting} as baselines for adversarial training and distillation.
However, our training approach remains independent of the choice of the training and distillation methods.
We design dynamic networks using bottleneck residual blocks and inverted bottleneck blocks to support both ResNet and MobileNet architectures.
In this work, the dynamic ResNet network is parameterized by Depth (D), Width (W), and Expansion (E), while for the dynamic MobileNet network, we replace the width with Kernel size(K) in its configuration parameters.
The experiments are conducted on the CIFAR-10 and CIFAR-100 datasets.
\subsubsection{Training and Evaluation Details}
To train the dynamic network, we first adversarially train the largest dynamic teacher network using the TRADES~\cite{zhang2019theoretically} method for $300$ epochs.
Next, we allocate $120$ epochs for each step of training dynamic width (kernel size), depth, and expansion for the student networks, sharing the weights with the dynamic network.
The training process uses the stochastic gradient descent (SGD) optimizer with an initial learning rate of $0.01$, a momentum of $0.9$, and a weight decay of $2e-4$. The batch size is set to $128$.
For the PGD attack used during TRADES training, we adopt the $l_{\infty}$ norm with $10$ iterations, a step size of $2/255$, and $\epsilon=8/255$.  
In ProARD, we consider a list of values for one configuration parameter (e.g., depth, width (kernel size), and expansion) at each stage of the dynamic network.
For instance, we consider $W=\{0.65,0.8,1.0\}$, $D=\{0,1,2\}$, and $E=\{0.2,0.25,0.35\}$ for a dynamic ResNet network.
For the dynamic MobileNet network, we used $K=\{3,5,7\}$ instead of width in the stage design.
Our method is agnostic to the choice of loss function used in robustness distillation. For this study, we employ RSLAD~\cite{zi2021revisiting}, which leverages robust soft-label adversarial distillation.
After training the dynamic network, we evaluate each student's architecture against $2$ adversarial attacks:  FGSM, PGD20. 
Maximum perturbation used for evaluation is also set to $\epsilon=8/255$ and $20$ perturbation iterations.
\begin{table}[bp]
\centering
\captionsetup{justification=centering}
\resizebox{\columnwidth}{!}{
\begin{tabular}{cc}
        \begin{tikzpicture}
        \begin{axis}[
            ybar,
            font=\Large,
            x tick label style = {text width=2cm,align=center},
            enlargelimits=0.20,
            legend style={at={(0.5,1.15)},
              anchor=north,legend columns=-1},
            ylabel=\textbf{Accuracy},
            ybar=8pt,
            symbolic x coords={D=2 E=0.35 W=0.65,D=2 E=0.2 W=1,D=1 E=0.35 W=1},
            xtick=data,
            nodes near coords,
            nodes near coords align={vertical},
            ]
        \addplot[draw=blue,fill=blue] coordinates {(D=2 E=0.35 W=0.65,85.2) (D=2 E=0.2 W=1,84) (D=1 E=0.35 W=1,84.3) };
        \addplot [draw=red ,fill=red]coordinates {(D=2 E=0.35 W=0.65,70.2) (D=2 E=0.2 W=1,70.1) (D=1 E=0.35 W=1,70.7)};
        \coordinate (A) at (axis cs:D=2 E=0.35 W=0.65,85.2);
        \coordinate (O1) at (rel axis cs:0,0);
        \coordinate (O2) at (rel axis cs:1,0);
        \draw [green,sharp plot,dashed] (A -| O1) -- (A -| O2);
        \legend{ProARD,Random}
        \end{axis}
        \end{tikzpicture}
        &
       \begin{tikzpicture}
        \begin{axis}[
            ybar,
            font=\Large,
            x tick label style = {text width=2cm,align=center},
            enlargelimits=0.20,
            legend style={at={(0.5,1.15)},
              anchor=north,legend columns=-1},
            ylabel=\textbf{Robustness},
            ybar=8pt,
            symbolic x coords={D=2 E=0.35 W=0.65,D=2 E=0.2 W=1,D=1 E=0.35 W=1},
            xtick=data,
            nodes near coords,
            nodes near coords align={vertical},
            ]
        \addplot[draw=blue,fill=blue] coordinates {(D=2 E=0.35 W=0.65,53.04) (D=2 E=0.2 W=1,50.8) (D=1 E=0.35 W=1,50.9) };
        \addplot [draw=red ,fill=red]coordinates {(D=2 E=0.35 W=0.65,39.3) (D=2 E=0.2 W=1,38.8) (D=1 E=0.35 W=1,39.6)};
        \coordinate (A) at (axis cs:D=2 E=0.35 W=0.65,54.3);
        \coordinate (O1) at (rel axis cs:0,0);
        \coordinate (O2) at (rel axis cs:1,0);
        \draw [green,sharp plot,dashed] (A -| O1) -- (A -| O2);
        \legend{ProARD,Random}
        \end{axis}
        \end{tikzpicture}
\end{tabular}}

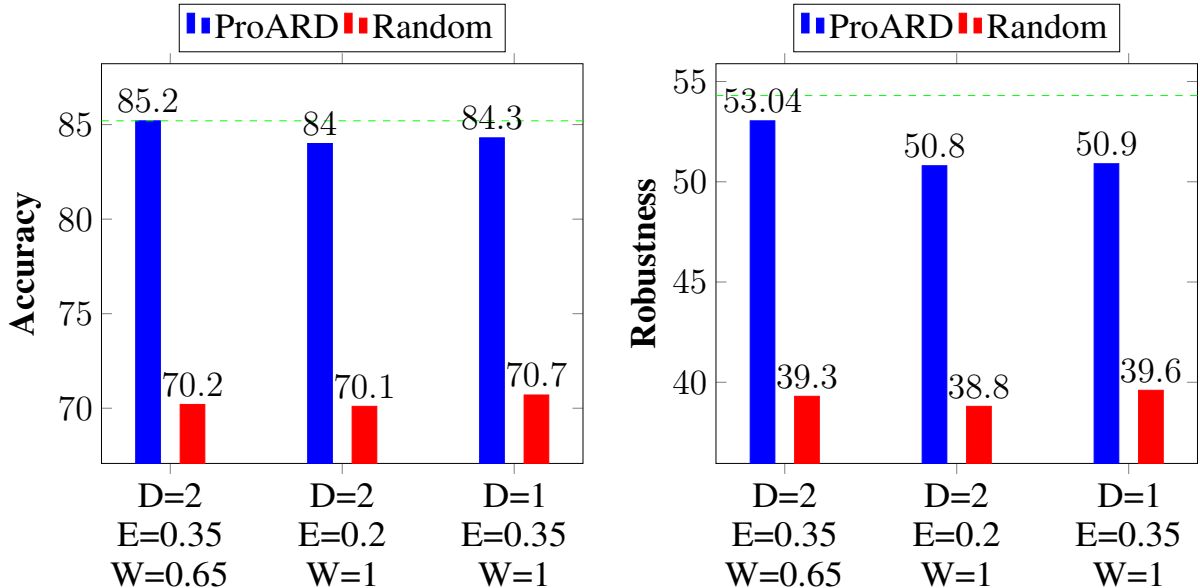
\captionof{figure}{(Left) Accuracy and (Right) Robustness for three students extracted from the Dynamic ResNet on the CIFAR-10 dataset. The green line shows the performance of largest student network architecture trained with TRADES~\cite{zhang2019theoretically}} 
\label{fig:students}
\vspace{-0.5 cm}
\end{table}
\subsubsection{Multi-objective Search}

For the multi-objective search, we use the NSGA-II evolutionary algorithm, which leverages non-dominated sorting and crowding distance mechanisms to tackle the optimization problem.
The configuration parameters in each stage and layer of each architecture are encoded as a genotype representation.
The objective function combines predicted robustness and predicted accuracy, while the search is constrained by the number of FLOPs.
Given the computational complexity of directly evaluating robustness and accuracy, we develop an accuracy-robustness predictor based on a simple fully-connected neural network.
To train the predictor, we generate a dataset comprising 2,000 samples of diverse network architectures and compute their robustness and accuracy under specific adversarial attack settings (PGD ($\epsilon=0.031, iter = 20, step-size=0.0071$).
The predictor is trained to efficiently estimate the robustness and accuracy of any given architecture configuration.
Using NSGA-II, we identify the optimal student network architecture that strikes the best balance between robustness and accuracy, subject to a specified number of FLOPs constraint.

\subsection{Random sampling vs Progressive}
\begin{figure}[bp]
\centering
\resizebox{\columnwidth}{!}{
\begin{tikzpicture}
  \begin{axis}[
    xbar,
    width=12cm, height=3.5cm, enlarge y limits=0.5,
    xlabel={GPU Hours},
    symbolic y coords={ProARD, TRADES, RSLAD},
    ytick=data,
    nodes near coords, nodes near coords align={horizontal},
  ]
    \addplot [fill=cyan] coordinates {(15,ProARD) (500,TRADES) (900,RSLAD)};
    
    \draw[red,ultra thick, <->] (axis cs:80,ProARD) -- (axis cs:900,ProARD) node[midway, below] {\textbf{60x}};

  \end{axis}
\end{tikzpicture}}
\caption{Total Cost (GPU Hours) For 50 edge devices}
\label{fig1:cost}
\end{figure}
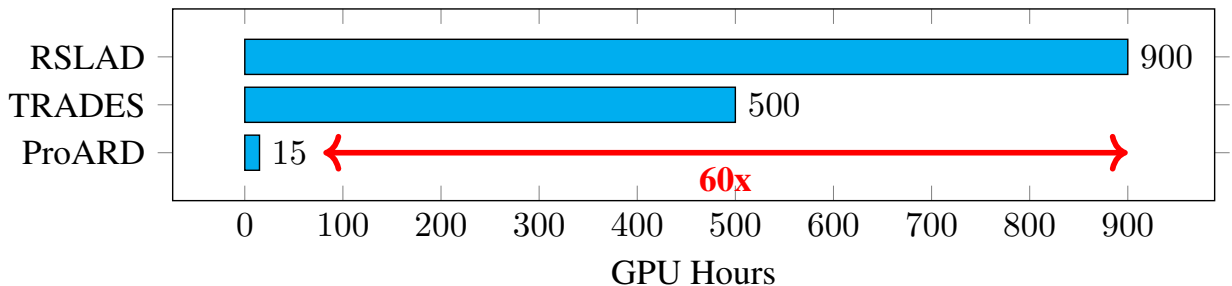

We compare the random sampling robustness distillation approach that randomly selects some student networks and solves the optimization problem(Eq.~\ref{adv_dyn}) to train the dynamic network with our ProARD method.
In Fig~\ref{fig:studentDistribution} we plot the accuracy-robustness distribution of $1000$ different student architectures extracted from Dynamic ResNet and Dynamic MobileNet on the CIFAR-10 dataset after training.
ProARD significantly improves the accuracy and robustness of the student networks in comparison with random sampling. 
ProARD achieves a $13\%$ and $10\%$ accuracy improvement by selecting the most accurate student and a $14\%$ and $5\%$ robustness improvement when considering the most robust student network in both dynamic networks.
\begin{table*}[htbp]
\centering
\captionsetup{justification=centering}
\resizebox{\columnwidth}{!}{
\begin{tabular}{cc|cc}
        \begin{tikzpicture}
     \begin{axis}[
            width=\columnwidth,
            font=\Large,
            scaled x ticks = true,
            title =  \textbf{Random Sampling (Dynamic ResNet)},
            scaled y ticks = true,
            ymin=37, ymax=42,
            xmin =69, xmax=72,
            grid=major, 
            grid style={dashed,gray!40}, 
            ylabel near ticks,
            xlabel near ticks,
            xlabel= \textbf{Accuracy(\%)}, 
            ylabel= \textbf{Robustness(\%)},
            xticklabel style={yshift=-1ex}, 
            yticklabel style={xshift=-1ex}, 
            xlabel style={yshift=-1ex},    
            ylabel style={xshift=-1.5ex},   
            ]
            \addplot [red,only marks] table[x=acc, y=rob, col sep=comma] {wps_csv_resnet50.csv};
        \end{axis}
        \end{tikzpicture}
  &
       \begin{tikzpicture}
     \begin{axis}[
            width=\columnwidth,
            font=\Large,
            title =  \textbf{ProARD (Dynamic ResNet)},
            scaled x ticks = true,
            scaled y ticks = true,
            ymin=44, ymax=55,
            xmin =82, xmax=86,
            grid=major, 
            grid style={dashed,gray!40}, 
            ylabel near ticks,
            xlabel near ticks,
            xlabel= \textbf{Accuracy(\%)}, 
            ylabel= \textbf{Robustness(\%)},
            xticklabel style={yshift=-1ex}, 
            yticklabel style={xshift=-1ex}, 
            xlabel style={yshift=-1ex},    
            ylabel style={xshift=-1.5ex},   
            ]
            \addplot [blue,only marks] table[x=acc, y=rob, col sep=comma] {csv.csv};
        \end{axis}
        \end{tikzpicture}
        &
        \begin{tikzpicture}
     \begin{axis}[
            width=\columnwidth,
            font=\Large,
            title =  \textbf{Random Sampling (Dynamic MobileNet)},
            scaled x ticks = true,
            scaled y ticks = true,
            ymin=38, ymax=44,
            xmin =70, xmax=73,
            grid=major, 
            grid style={dashed,gray!40}, 
            ylabel near ticks,
            xlabel near ticks,
            xlabel= \textbf{Accuracy(\%)}, 
            ylabel= \textbf{Robustness(\%)},
            xticklabel style={yshift=-1ex}, 
            yticklabel style={xshift=-1ex}, 
            xlabel style={yshift=-1ex},    
            ylabel style={xshift=-1.5ex},   
            ]
            \addplot [red,only marks] table[x=acc, y=rob, col sep=comma] {acc_rob_mbv3_wps.csv};
        \end{axis}
        \end{tikzpicture}
        &
        \begin{tikzpicture}
     \begin{axis}[
            width=\columnwidth,
            font=\Large,
            scaled x ticks = true,
            scaled y ticks = true,
            title =  \textbf{ProARD (Dynamic MobileNet)},
            ymin=37, ymax=50,
            xmin =75, xmax=81,
            grid=major, 
            grid style={dashed,gray!40}, 
            ylabel near ticks,
            xlabel near ticks,
            xlabel= \textbf{Accuracy(\%)}, 
            ylabel= \textbf{Robustness(\%)},
            xticklabel style={yshift=-1ex}, 
            yticklabel style={xshift=-1ex}, 
            xlabel style={yshift=-1ex},    
            ylabel style={xshift=-1.5ex},   
            ]
            \addplot [blue,only marks] table[x=acc, y=rob, col sep=comma] {acc_rob_mbv3.csv};
        \end{axis}
        \end{tikzpicture}
\end{tabular}}

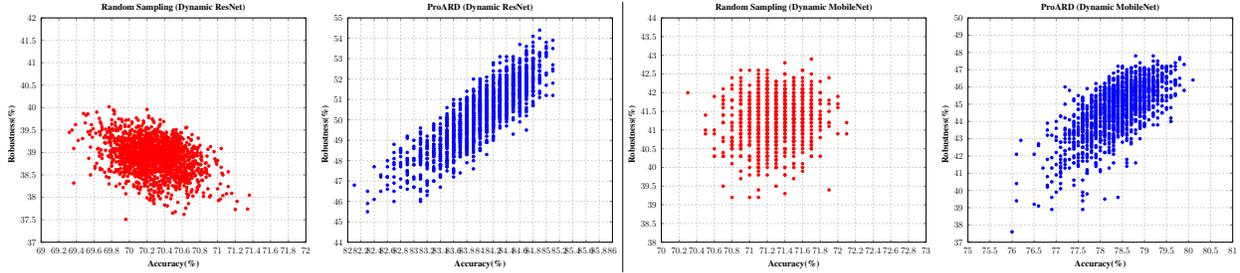
\captionof{figure}{The accuracy-robustness distribution of students for dynamic networks with random sampling and ProARD on CIFAR-10 dataset: (Left) Dynamic ResNet, (Right) Dynamic MobileNet.} 
\label{fig:studentDistribution}
\vspace{-0.5 cm}
\end{table*}
Fig~\ref{fig:students} reports the accuracy and robustness of student networks extracted from Dynamic ResNet trained on the CIFAR-10 dataset with ProARD and random sampling. 
Due to space limitations, we select $3$ students for comparison by varying the width of the dynamic ResNet with fixed depth and expansion.
ProARD yields better accuracy and robustness for the selected students compared to random sampling.
For example, the student with $D=2, E=0.35,\text{and }  W=0.65$ delivers nearly the same accuracy and robustness as the largest network (green line), but with fewer parameters.
Fig~\ref{fig1:cost} shows the total cost required for training student networks across 50 different deployment scenarios (edge devices).
ProARD reduces the total cost (GPU hours) by a factor of 60 compared to RSLAD by training a dynamic network and efficiently extracting students for each device.

\begin{table}[hb]
\centering
\captionsetup{justification=centering}
\resizebox{\columnwidth}{!}{
\begin{tabular}{cc}
    \centering
    \begin{tikzpicture}
      \begin {axis}[
        width=0.8\columnwidth,
        style={at={(0.5,1.07)},anchor=north,yshift=-0.1},
        xmin=82, xmax=86,
        ymin=80.5, ymax=88,
        grid=major, 
        grid style={dashed,gray!40},
        xlabel= \textbf{Real Accuracy}, 
        ylabel= \textbf{Predicted Accuracy},
        legend style={at={(0.5,-0.2)},anchor=north, nodes={scale=1, transform shape}}, 
        legend pos=north west,
        label style={font=\small}, tick label style={font=\small} 
        ]
       \addplot [blue, only marks,   thin, mark options={scale=0.7}]
        table[x=acc,y=pacc,col sep=comma] {accs.csv}; 
        
        \addplot [ mark = none, red,  line width=0.5mm ] coordinates {(82,82) (86,86)};

        \node[draw] at (84.9,81.8) {RMSE = 0.0076};
      \end{axis}
    \end{tikzpicture}
&
    \begin{tikzpicture}
    
      \begin {axis}[
        width=0.8\columnwidth,
        style={at={(0.5,1.07)},anchor=north,yshift=-0.1},
        xmin=47, xmax=55,
        ymin=47, ymax=55,
        grid=major, 
        grid style={dashed,gray!40},
        xlabel= \textbf{Real Robustness}, 
        ylabel= \textbf{Predicted Robustness},
        legend style={at={(0.5,-0.2)},anchor=north, nodes={scale=1, transform shape}}, 
        legend pos=north west,
        label style={font=\small}, tick label style={font=\small} 
        ]
       \addplot [blue, only marks,   thin, mark options={scale=0.7}]
        table[x=rob,y=prob,col sep=comma] {robs.csv}; 
        
        \addplot [ mark = none, red,  line width=0.5mm ] coordinates {(47,47) (55,55)};
        
        \node[draw] at (52.8,48.3) {RMSE = 0.0072};
      \end{axis}
    \end{tikzpicture}
\end{tabular}}

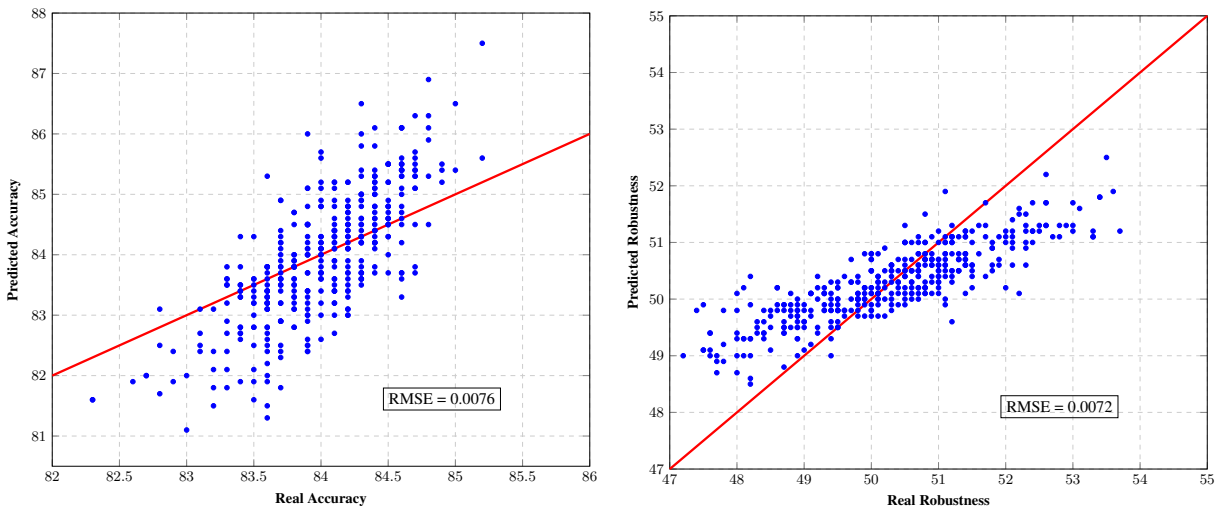
\captionof{figure}{Evaluation performance of the accuracy-robustness predictor on 300 different test student architectures from Dyanmic ResNet on CIFAR-10 dataset.} 
\label{fig:predictor}
\vspace{-0.5 cm}
\end{table}
\subsection{Accuracy-Robustness Predictor}
For the multi-objective evolutionary search, a fitness function is required.
However, evaluating accuracy and robustness for different architectures is time-consuming.
To address this, we train a prediction model based on a fully connected network to predict the accuracy and robustness of each architecture.
To create a dataset for training, we extract 2000 student networks from the dynamic network, generate a feature vector based on the width (kernel), depth, and expansion of each stage, and use this vector as the features of the architecture.
We then evaluate the accuracy and robustness of these networks, using them as labels for training.
The fully connected network is trained for $30$ epochs.
We compare the accuracy of our predictor across $300$ different architectures, evaluating the actual accuracy and robustness of each and comparing them with the predicted values.
Fig~\ref{fig:predictor} shows the difference between the real accuracy and robustness and the predicted values.
Our predictor can estimate accuracy and robustness for test architectures with a root mean squared error (RMSE) of $0.0076$ for accuracy and $0.0072$ for robustness.

\subsection{Search student networks}
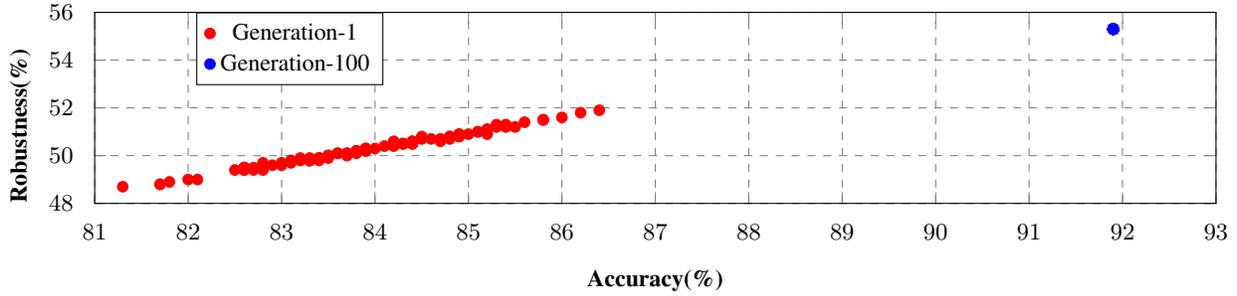
\begin{figure}[htbp]
 \resizebox{\columnwidth}{!}{
  \begin{tikzpicture}[spy using outlines=
	{circle, magnification=2, connect spies}]
    
     \begin{axis}[
            width=\columnwidth,
            height = 0.25\textwidth,
            font=\small,
            legend style={at={(0.174,1)},
              anchor=north,legend columns=1, font=\small},
            scaled x ticks = true,
            scaled y ticks = true,
            ymin=48, ymax=56,
            xmin =81, xmax=93,
            grid=major, 
            grid style={dashed,gray}, 
            ylabel near ticks,
            xlabel near ticks,
            xlabel= \textbf{Accuracy(\%)}, 
            ylabel= \textbf{Robustness(\%)},
            xticklabel style={yshift=-1ex}, 
            yticklabel style={xshift=-1ex}, 
            xlabel style={yshift=-1ex},    
            ylabel style={xshift=-1.5ex},   
            ]

            \addplot [red,only marks] table[x=acc0, y=rob0, col sep=comma] {acc_rob_gen.csv};
             \addplot [blue,only marks] table[x=acc99, y=rob99, col sep=comma] {acc_rob_gen.csv};
            \legend{Generation-1, Generation-100}
        \end{axis}
        \end{tikzpicture}}
       
\caption{Representation of the first-generation and final student networks found by NSGA-II, based on our accuracy-robustness predictor, for a fixed FLOPs constraint in the dynamic ResNet on the CIFAR-10 dataset.}
\label{fig1:RD_M}
\end{figure}

To find the best architecture based on accuracy, robustness, and a given FLOPs constraint, we use NSGA-II as the multi-objective search framework.
We begin with a random initial population and perform $100$ iterations with a mutation rate of $0.1$ to identify the optimal student architecture.
To accelerate the search process, we use the predictor trained in the previous step to evaluate all individuals in the population.
Fig~\ref{fig1:RD_M} shows the accuracy and robustness of the first population and the final student networks after $100$ generations.
NSGA-II successfully identifies the best architecture, achieving better robustness and accuracy while maintaining the same number of FLOPs.
In this case, we set our FLOPs constraint to be less than $2000$.
We choose FLOPs as a metric because it is independent of hardware-specific architectural details and effectively reflects how well a system can leverage parallelism, a fundamental characteristic of modern accelerators.

\subsection{White-box Attacks}

We evaluate the accuracy and robustness of the best student network identified through the multi-objective search for white-box attacks and present the results in Table~\ref{Cifar10-whitebox-attacks} for CIFAR-10 and CIFAR-100 datasets.
We report the best robustness validated by FGSM and PGD-20 attack methods.
We utilize Dynamic ResNet and Dynamic MobileNet as the dynamic architectures. 
\begin{table*}[h!]
    \centering
    \caption{Accuracy and white-box robustness results on CIFAR-10 and CIFAR-100 datasets. The best results are \textbf{blodfaced}. For RSLAD, we used WideResNet-34-10 as the teacher network}   
    \label{Cifar10-whitebox-attacks}
    \resizebox{0.9\linewidth}{!}{%
    \begin{tabular}{c|c|c|cccc|c}
    \toprule[1pt]
        \textbf{Dataset}&\textbf{Networks} &  \textbf{\#Params(M)}&  \textbf{Methods} & \textbf{Natural Acc.}& \textbf{FGSM}& \textbf{PGD20} & \textbf{Training} \\ \midrule
        \multirow{7}{*}{CIFAR10}& ResNet50 & 23.52& TRADES & 84.7 & 59.7 &  52.6 & Yes (train from scratch)  \\
        & MBV3 & 4.21 &TRADES & 80.0 & 52.2 & 48.6 & Yes (train from scratch) \\
        & ResNet50 (WideResNet) &23.52& RSLAD & 86.0 & 59.4 & 52.6 & Yes (train from scratch) \\
        & MBV3 (WideResNet)& 4.21& RSLAD & \textbf{81.2} & 54.5 & 50.5 & Yes (train from scratch)  \\
        & \textbf{Dyn-Resnet50} & 24.94& \textbf{ProARD} &\textbf{87.0}& \textbf{62.8}  &\textbf{54.2} & \textbf{No (quick search)}  \\
        & \textbf{Dyn-MBV3} &5.28&\textbf{ProARD} & 80.6 & \textbf{55.0} & \textbf{50.9} & \textbf{No (quick search)}\\
        \midrule[1.1pt] 
        \multirow{7}{*}{CIFAR100}& ResNet50 & 23.71 & TRADES & 54.6 & 28.7 & 25.8 & Yes (train from scratch)  \\
        & MBV3 & 4.33 &TRADES & 53.4 & 28.9  & 27.2 & Yes (train from scratch) \\
        & ResNet50 (WideResNet) &23.71& RSLAD & 55.3 & \textbf{29.6} & \textbf{26.5}  & Yes (train from scratch) \\
        & MBV3 (WideResNet)&4.33 & RSLAD & 54.9   & \textbf{29.1} & \textbf{28.7} & Yes (train from scratch) \\
        & \textbf{Dyn-Resnet50} &27.24& \textbf{ProARD} & \textbf{60.1} & 29.4 & 25.9 & \textbf{No (quick search)} \\
        & \textbf{Dyn-MBV3} & 5.41 & \textbf{ProARD} & \textbf{55.3}  &28.6 &27.8 & \textbf{No (quick search)} \\
        \midrule[1.1pt]
    \end{tabular}}
\end{table*}
These results suggest that ProARD delivers equal or superior accuracy and robustness for student networks with the same number of FLOPs, without requiring retraining, by simply performing a quick search within the dynamic network. ProARD identifies models with a higher number of parameters while maintaining the same FLOPs, effectively increasing model capacity. This aligns with previous research suggesting that increasing network capacity can enhance robustness~\cite{bubeck2021universal}.

\section{Conclusion}
In this paper, we address the challenge of training a vast number of student networks (over $10^{19}$) for robustness distillation.
We introduce a dynamic network that supports diverse configurations, including variations in depth, width (kernel size), and expansion. 
Training each student individually is computationally infeasible, and random sampling combined with weight sharing fails to produce accurate and robust networks.
To overcome these limitations, we propose Progressive Adversarial Robustness Distillation (ProARD), a novel approach that efficiently trains a dynamic network to generate robust student networks without retraining.
ProARD employs a progressive sampling during training, coupled with a multi-objective search powered by an accuracy-robustness predictor, to quickly identify optimal architectures tailored to specific resource constraints.
Our results demonstrate that ProARD reduces training costs while delivering students with higher accuracy and improved robustness compared to random sampling. 
ProARD offers an efficient framework for training robust dynamic networks, enabling the rapid selection of optimized architectures for various hardware constraints without the need for retraining.
This significantly reduces computational costs and supports scalable deployment across diverse devices.

\section*{Acknowledgment}
This work was partly supported by the European Union and the Estonian Research Council via project TEM-TA138, the Swedish Innovation Agency VINNOVA projects AutoDeep and FASTER-AI. “The computations were enabled by resources provided by the National Academic Infrastructure for Supercomputing in Sweden (NAISS), funded by the Swedish Research Council through grant agreements 2022-06725 and 2024-221034.

\bibliographystyle{unsrt}

\begin{thebibliography}{88}


\ifx \showCODEN    \undefined \def \showCODEN     #1{\unskip}     \fi
\ifx \showDOI      \undefined \def \showDOI       #1{#1}\fi
\ifx \showISBNx    \undefined \def \showISBNx     #1{\unskip}     \fi
\ifx \showISBNxiii \undefined \def \showISBNxiii  #1{\unskip}     \fi
\ifx \showISSN     \undefined \def \showISSN      #1{\unskip}     \fi
\ifx \showLCCN     \undefined \def \showLCCN      #1{\unskip}     \fi
\ifx \shownote     \undefined \def \shownote      #1{#1}          \fi
\ifx \showarticletitle \undefined \def \showarticletitle #1{#1}   \fi
\ifx \showURL      \undefined \def \showURL       {\relax}        \fi
\providecommand\bibfield[2]{#2}
\providecommand\bibinfo[2]{#2}
\providecommand\natexlab[1]{#1}
\providecommand\showeprint[2][]{arXiv:#2}
\bibitem{goodfellow2014explaining}Goodfellow, I., Shlens, J. \& Szegedy, C. Explaining and harnessing adversarial examples. {\em ArXiv Preprint ArXiv:1412.6572}. (2014)
\bibitem{kannan2018adversarial}Kannan, H., Kurakin, A. \& Goodfellow, I. Adversarial logit pairing. {\em ArXiv Preprint ArXiv:1803.06373}. (2018)
\bibitem{madry2017towards}Madry, A., Makelov, A., Schmidt, L., Tsipras, D. \& Vladu, A. Towards deep learning models resistant to adversarial attacks. {\em ArXiv Preprint ArXiv:1706.06083}. (2017)
\bibitem{zhang2019theoretically}Zhang, H., Yu, Y., Jiao, J., Xing, E., El Ghaoui, L. \& Jordan, M. Theoretically principled trade-off between robustness and accuracy. {\em International Conference On Machine Learning}. pp. 7472-7482 (2019)
\bibitem{cui2021learnable}Cui, J., Liu, S., Wang, L. \& Jia, J. Learnable boundary guided adversarial training. {\em Proceedings Of The IEEE/CVF International Conference On Computer Vision}. pp. 15721-15730 (2021)
\bibitem{zhang2020attacks}Zhang, J., Xu, X., Han, B., Niu, G., Cui, L., Sugiyama, M. \& Kankanhalli, M. Attacks which do not kill training make adversarial learning stronger. {\em International Conference On Machine Learning}. pp. 11278-11287 (2020)
\bibitem{xie2019intriguing}Xie, C. \& Yuille, A. Intriguing properties of adversarial training at scale. {\em ArXiv Preprint ArXiv:1906.03787}. (2019)
\bibitem{huang2021exploring}Huang, H., Wang, Y., Erfani, S., Gu, Q., Bailey, J. \& Ma, X. Exploring architectural ingredients of adversarially robust deep neural networks. {\em Advances In Neural Information Processing Systems}. \textbf{34} pp. 5545-5559 (2021)
\bibitem{rice2020overfitting}Rice, L., Wong, E. \& Kolter, Z. Overfitting in adversarially robust deep learning. {\em International Conference On Machine Learning}. pp. 8093-8104 (2020)
\bibitem{cai2021network}Cai, H., Gan, C., Lin, J. \& Han, S. Network augmentation for tiny deep learning. {\em ArXiv Preprint ArXiv:2110.08890}. (2021)
\bibitem{zi2021revisiting}Zi, B., Zhao, S., Ma, X. \& Jiang, Y. Revisiting adversarial robustness distillation: Robust soft labels make student better. {\em Proceedings Of The IEEE/CVF International Conference On Computer Vision}. pp. 16443-16452 (2021)
\bibitem{carlini2017towards}Carlini, N. \& Wagner, D. Towards evaluating the robustness of neural networks. {\em 2017 Ieee Symposium On Security And Privacy (sp)}. pp. 39-57 (2017)
\bibitem{croce2020reliable}Croce, F. \& Hein, M. Reliable evaluation of adversarial robustness with an ensemble of diverse parameter-free attacks. {\em International Conference On Machine Learning}. pp. 2206-2216 (2020)
\bibitem{wang2020improving}Wang, Y., Zou, D., Yi, J., Bailey, J., Ma, X. \& Gu, Q. Improving adversarial robustness requires revisiting misclassified examples. {\em International Conference On Learning Representations}. (2020)
\bibitem{zhang2020geometry}Zhang, J., Zhu, J., Niu, G., Han, B., Sugiyama, M. \& Kankanhalli, M. Geometry-aware instance-reweighted adversarial training. {\em ArXiv Preprint ArXiv:2010.01736}. (2020)
\bibitem{yu2022understanding}Yu, C., Han, B., Shen, L., Yu, J., Gong, C., Gong, M. \& Liu, T. Understanding robust overfitting of adversarial training and beyond. {\em International Conference On Machine Learning}. pp. 25595-25610 (2022)
\bibitem{costa2024deep}Costa, J., Roxo, T., Proença, H.  Inácio, P. How deep learning sees the world: A survey on adversarial attacks \& defenses. {\em IEEE Access}. (2024)
\bibitem{rade2021helper}Rade, R. \& Moosavi-Dezfooli, S. Helper-based adversarial training: Reducing excessive margin to achieve a better accuracy vs. robustness trade-off. {\em ICML 2021 Workshop On Adversarial Machine Learning}. (2021)
\bibitem{shafahi2019adversarial}Shafahi, A., Najibi, M., Ghiasi, M., Xu, Z., Dickerson, J., Studer, C., Davis, L., Taylor, G. \& Goldstein, T. Adversarial training for free!. {\em Advances In Neural Information Processing Systems}. \textbf{32} (2019)
\bibitem{wong2020fast}Wong, E., Rice, L. \& Kolter, J. Fast is better than free: Revisiting adversarial training. {\em ArXiv Preprint ArXiv:2001.03994}. (2020)
\bibitem{goldblum2020adversarially}Goldblum, M., Fowl, L., Feizi, S. \& Goldstein, T. Adversarially robust distillation. {\em Proceedings Of The AAAI Conference On Artificial Intelligence}. \textbf{34}, 3996-4003 (2020)
\bibitem{zhu2021reliable}Zhu, J., Yao, J., Han, B., Zhang, J., Liu, T., Niu, G., Zhou, J., Xu, J. \& Yang, H. Reliable adversarial distillation with unreliable teachers. {\em ArXiv Preprint ArXiv:2106.04928}. (2021)
\bibitem{zoph2018learning}Zoph, B., Vasudevan, V., Shlens, J. \& Le, Q. Learning transferable architectures for scalable image recognition. {\em Proceedings Of The IEEE Conference On Computer Vision And Pattern Recognition}. pp. 8697-8710 (2018)
\bibitem{zoph2016neural}Zoph, B. Neural architecture search with reinforcement learning. {\em ArXiv Preprint ArXiv:1611.01578}. (2016)
\bibitem{liu2018darts}Liu, H., Simonyan, K. \& Yang, Y. Darts: Differentiable architecture search. {\em ArXiv Preprint ArXiv:1806.09055}. (2018)
\bibitem{chitty2022neural}Chitty-Venkata, K. \& Somani, A. Neural architecture search survey: A hardware perspective. {\em ACM Computing Surveys}. \textbf{55}, 1-36 (2022)
\bibitem{cai2018proxylessnas}Cai, H., Zhu, L. \& Han, S. Proxylessnas: Direct neural architecture search on target task and hardware. {\em ArXiv Preprint ArXiv:1812.00332}. (2018)
\bibitem{chitty2023neural}Chitty-Venkata, K., Emani, M., Vishwanath, V. \& Somani, A. Neural architecture search benchmarks: Insights and survey. {\em IEEE Access}. \textbf{11} pp. 25217-25236 (2023)
\bibitem{mao2021differentiable}Mao, Y., Zhong, G., Wang, Y. \& Deng, Z. Differentiable light-weight architecture search. {\em 2021 IEEE International Conference On Multimedia And Expo (ICME)}. pp. 1-6 (2021)
\bibitem{cai2019once}Cai, H., Gan, C., Wang, T., Zhang, Z. \& Han, S. Once-for-all: Train one network and specialize it for efficient deployment. {\em ArXiv Preprint ArXiv:1908.09791}. (2019)
\bibitem{sahni2021compofa}Sahni, M., Varshini, S., Khare, A. \& Tumanov, A. CompOFA: Compound once-for-all networks for faster multi-platform deployment. {\em ArXiv Preprint ArXiv:2104.12642}. (2021)
\bibitem{girard2024memory}Girard, M., Quétu, V., Tardieu, S., Nguyen, V. \& Tartaglione, E. Memory-Optimized Once-For-All Network. {\em ArXiv Preprint ArXiv:2409.05900}. (2024)
\bibitem{sakuma2023detofa}Sakuma, Y., Ishii, M. \& Narihira, T. DetOFA: Efficient Training of Once-for-All Networks for Object Detection using Path Filter. {\em Proceedings Of The IEEE/CVF International Conference On Computer Vision}. pp. 1333-1342 (2023)
\bibitem{wang2020hat}Wang, H., Wu, Z., Liu, Z., Cai, H., Zhu, L., Gan, C. \& Han, S. Hat: Hardware-aware transformers for efficient natural language processing. {\em ArXiv Preprint ArXiv:2005.14187}. (2020)
\bibitem{tang2020searching}Tang, H., Liu, Z., Zhao, S., Lin, Y., Lin, J., Wang, H. \& Han, S. Searching efficient 3d architectures with sparse point-voxel convolution. {\em European Conference On Computer Vision}. pp. 685-702 (2020)
\bibitem{lin2021anycost}Lin, J., Zhang, R., Ganz, F., Han, S. \& Zhu, J. Anycost gans for interactive image synthesis and editing. {\em Proceedings Of The IEEE/CVF Conference On Computer Vision And Pattern Recognition}. pp. 14986-14996 (2021)
\bibitem{wang2022lite}Wang, Y., Li, M., Cai, H., Chen, W. \& Han, S. Lite pose: Efficient architecture design for 2d human pose estimation. {\em Proceedings Of The IEEE/CVF Conference On Computer Vision And Pattern Recognition}. pp. 13126-13136 (2022)
\bibitem{krizhevsky2012imagenet}Krizhevsky, A., Sutskever, I. \& Hinton, G. Imagenet classification with deep convolutional neural networks. {\em Advances In Neural Information Processing Systems}. \textbf{25} (2012)
\bibitem{zou2023object}Zou, Z., Chen, K., Shi, Z., Guo, Y. \& Ye, J. Object detection in 20 years: A survey. {\em Proceedings Of The IEEE}. \textbf{111}, 257-276 (2023)
\bibitem{zou2024segment}Zou, X., Yang, J., Zhang, H., Li, F., Li, L., Wang, J., Wang, L., Gao, J. \& Lee, Y. Segment everything everywhere all at once. {\em Advances In Neural Information Processing Systems}. \textbf{36} (2024)
\bibitem{eykholt2018robust}Eykholt, K., Evtimov, I., Fernandes, E., Li, B., Rahmati, A., Xiao, C., Prakash, A., Kohno, T. \& Song, D. Robust physical-world attacks on deep learning visual classification. {\em Proceedings Of The IEEE Conference On Computer Vision And Pattern Recognition}. pp. 1625-1634 (2018)
\bibitem{sharif2016accessorize}Sharif, M., Bhagavatula, S., Bauer, L. \& Reiter, M. Accessorize to a crime: Real and stealthy attacks on state-of-the-art face recognition. {\em Proceedings Of The 2016 Acm Sigsac Conference On Computer And Communications Security}. pp. 1528-1540 (2016)
\bibitem{szegedy2013intriguing}Szegedy, C. Intriguing properties of neural networks. {\em ArXiv Preprint ArXiv:1312.6199}. (2013)
\bibitem{ou2024towards}Ou, Y., Feng, Y. \& Sun, Y. Towards Accurate and Robust Architectures via Neural Architecture Search. {\em Proceedings Of The IEEE/CVF Conference On Computer Vision And Pattern Recognition}. pp. 5967-5976 (2024)
\bibitem{mok2021advrush}Mok, J., Na, B., Choe, H. \& Yoon, S. AdvRush: Searching for adversarially robust neural architectures. {\em Proceedings Of The IEEE/CVF International Conference On Computer Vision}. pp. 12322-12332 (2021)
\bibitem{hosseini2021dsrna}Hosseini, R., Yang, X. \& Xie, P. Dsrna: Differentiable search of robust neural architectures. {\em Proceedings Of The IEEE/CVF Conference On Computer Vision And Pattern Recognition}. pp. 6196-6205 (2021)
\bibitem{yue2022effective}Yue, Z., Lin, B., Zhang, Y. \& Liang, C. Effective, efficient and robust neural architecture search. {\em 2022 International Joint Conference On Neural Networks (IJCNN)}. pp. 1-8 (2022)
\bibitem{cheng2023neural}Cheng, Z., Li, Y., Dong, M., Su, X., You, S. \& Xu, C. Neural architecture search for wide spectrum adversarial robustness. {\em Proceedings Of The AAAI Conference On Artificial Intelligence}. \textbf{37}, 442-451 (2023)
\bibitem{ou2022differentiable}Ou, Y., Xie, X., Gao, S., Sun, Y., Tan, K. \& Lv, J. Differentiable search of accurate and robust architectures. {\em ArXiv Preprint ArXiv:2212.14049}. (2022)
\bibitem{li2021neural}Li, Y., Yang, Z., Wang, Y. \& Xu, C. Neural architecture dilation for adversarial robustness. {\em Advances In Neural Information Processing Systems}. \textbf{34} pp. 29578-29589 (2021)
\bibitem{guo2020meets}Guo, M., Yang, Y., Xu, R., Liu, Z. \& Lin, D. When nas meets robustness: In search of robust architectures against adversarial attacks. {\em Proceedings Of The IEEE/CVF Conference On Computer Vision And Pattern Recognition}. pp. 631-640 (2020)
\bibitem{xie2023tiny}Xie, G., Wang, J., Yu, G., Lyu, J., Zheng, F. \& Jin, Y. Tiny adversarial multi-objective one-shot neural architecture search. {\em Complex \& Intelligent Systems}. \textbf{9}, 6117-6138 (2023)
\bibitem{feng2024lrnas}Feng, Y., Lv, Z., Chen, H., Gao, S., An, F. \& Sun, Y. LRNAS: Differentiable Searching for Adversarially Robust Lightweight Neural Architecture. {\em IEEE Transactions On Neural Networks And Learning Systems}. (2024)
\bibitem{wu2020skip}Wu, D., Wang, Y., Xia, S., Bailey, J. \& Ma, X. Skip connections matter: On the transferability of adversarial examples generated with resnets. {\em ArXiv Preprint ArXiv:2002.05990}. (2020)
\bibitem{shao2021adversarial}Shao, R., Shi, Z., Yi, J., Chen, P. \& Hsieh, C. On the adversarial robustness of vision transformers. {\em ArXiv Preprint ArXiv:2103.15670}. (2021)
\bibitem{zhu2023robust}Zhu, X., Li, J., Liu, Y. \& Wang, W. Robust Neural Architecture Search. {\em ArXiv Preprint ArXiv:2304.02845}. (2023)
\bibitem{dong2020adversarially}Dong, M., Li, Y., Wang, Y. \& Xu, C. Adversarially robust neural architectures. {\em ArXiv Preprint ArXiv:2009.00902}. (2020)
\bibitem{chen2020anti}Chen, H., Zhang, B., Xue, S., Gong, X., Liu, H., Ji, R. \& Doermann, D. Anti-bandit neural architecture search for model defense. {\em Computer Vision–ECCV 2020: 16th European Conference, Glasgow, UK, August 23–28, 2020, Proceedings, Part XIII 16}. pp. 70-85 (2020)
\bibitem{du2021learning}Du, X., Zhang, J., Han, B., Liu, T., Rong, Y., Niu, G., Huang, J. \& Sugiyama, M. Learning diverse-structured networks for adversarial robustness. {\em International Conference On Machine Learning}. pp. 2880-2891 (2021)
\bibitem{ha2024generalizable}Ha, H., Kim, M. \& Hwang, S. Generalizable lightweight proxy for robust NAS against diverse perturbations. {\em Advances In Neural Information Processing Systems}. \textbf{36} (2024)
\bibitem{wu2024robust}Wu, Y., Liu, F., Simon-Gabriel, C., Chrysos, G. \& Cevher, V. Robust NAS under adversarial training: benchmark, theory, and beyond. {\em ArXiv Preprint ArXiv:2403.13134}. (2024)
\bibitem{jung2023neural}Jung, S., Lukasik, J. \& Keuper, M. Neural architecture design and robustness: A dataset. {\em ArXiv Preprint ArXiv:2306.06712}. (2023)
\bibitem{peng2023robarch}Peng, S., Xu, W., Cornelius, C., Li, K., Duggal, R., Chau, D. \& Martin, J. Robarch: Designing robust architectures against adversarial attacks. {\em ArXiv Preprint ArXiv:2301.03110}. (2023)
\bibitem{huang2023revisiting}Huang, S., Lu, Z., Deb, K. \& Boddeti, V. Revisiting residual networks for adversarial robustness. {\em Proceedings Of The IEEE/CVF Conference On Computer Vision And Pattern Recognition}. pp. 8202-8211 (2023)
\bibitem{yang2024hybrid}Yang, S., Sun, X., Xu, K., Liu, Y., Tian, Y. \& Zhang, X. Hybrid Architecture-Based Evolutionary Robust Neural Architecture Search. {\em IEEE Transactions On Emerging Topics In Computational Intelligence}. (2024)
\bibitem{feng2024lrnas}Feng, Y., Lv, Z., Chen, H., Gao, S., An, F. \& Sun, Y. LRNAS: Differentiable Searching for Adversarially Robust Lightweight Neural Architecture. {\em IEEE Transactions On Neural Networks And Learning Systems}. (2024)
\bibitem{zhao2022enhanced}Zhao, S., Yu, J., Sun, Z., Zhang, B. \& Wei, X. Enhanced accuracy and robustness via multi-teacher adversarial distillation. {\em European Conference On Computer Vision}. pp. 585-602 (2022)
\bibitem{huang2023boosting}Huang, B., Chen, M., Wang, Y., Lu, J., Cheng, M. \& Wang, W. Boosting accuracy and robustness of student models via adaptive adversarial distillation. {\em Proceedings Of The IEEE/CVF Conference On Computer Vision And Pattern Recognition}. pp. 24668-24677 (2023)
\bibitem{vaswani2017attention}Vaswani, A. Attention is all you need. {\em Advances In Neural Information Processing Systems}. (2017)
\bibitem{ma2021understanding}Ma, X., Niu, Y., Gu, L., Wang, Y., Zhao, Y., Bailey, J. \& Lu, F. Understanding adversarial attacks on deep learning based medical image analysis systems. {\em Pattern Recognition}. \textbf{110} pp. 107332 (2021)
\bibitem{jung2024peeraid}Jung, J., Jang, H., Song, J. \& Lee, J. PeerAiD: Improving Adversarial Distillation from a Specialized Peer Tutor. {\em Proceedings Of The IEEE/CVF Conference On Computer Vision And Pattern Recognition}. pp. 24482-24491 (2024)
\bibitem{takahashi2022ardir}Takahashi, T., Yamada, M., Yamanaka, Y. \& Yamashita, T. ARDIR: Improving Robustness using Knowledge Distillation of Internal Representation. {\em ArXiv Preprint ArXiv:2211.00239}. (2022)
\bibitem{zhou2024derd}Zhou, Y., Zhang, Y., Zhang, L. \& Hua, Z. DERD: data-free adversarial robustness distillation through self-adversarial teacher group. {\em Proceedings Of The 32nd ACM International Conference On Multimedia}. pp. 10055-10064 (2024)
\bibitem{yin2024adversarial}Yin, S., Xiao, Z., Song, M. \& Long, J. Adversarial Distillation Based on Slack Matching and Attribution Region Alignment. {\em Proceedings Of The IEEE/CVF Conference On Computer Vision And Pattern Recognition}. pp. 24605-24614 (2024)
\bibitem{deng2024distilling}Deng, J., Palmer, A., Mahmood, R., Rathbun, E., Bi, J., Mahmood, K. \& Aguiar, D. Distilling Adversarial Robustness Using Heterogeneous Teachers. {\em ArXiv Preprint ArXiv:2402.15586}. (2024)
\bibitem{shao2021and}Shao, R., Yi, J., Chen, P. \& Hsieh, C. How and when adversarial robustness transfers in knowledge distillation?. {\em ArXiv Preprint ArXiv:2110.12072}. (2021)
\bibitem{kuang2023improving}Kuang, H., Liu, H., Wu, Y., Satoh, S. \& Ji, R. Improving adversarial robustness via information bottleneck distillation. {\em Advances In Neural Information Processing Systems}. \textbf{36} pp. 10796-10813 (2023)
\bibitem{ham2024neo}Ham, S., Park, J., Han, D. \& Moon, J. NEO-KD: knowledge-distillation-based adversarial training for robust multi-exit neural networks. {\em Advances In Neural Information Processing Systems}. \textbf{36} (2024)
\bibitem{dong2024robust}Dong, J., Koniusz, P., Chen, J., Wang, Z. \& Ong, Y. Robust Distillation via Untargeted and Targeted Intermediate Adversarial Samples. {\em Proceedings Of The IEEE/CVF Conference On Computer Vision And Pattern Recognition}. pp. 28432-28442 (2024)
\bibitem{dong2025adversarially}Dong, J., Koniusz, P., Chen, J. \& Ong, Y. Adversarially Robust Distillation by Reducing the Student-Teacher Variance Gap. {\em European Conference On Computer Vision}. pp. 92-111 (2025)
\bibitem{zhou2024derd}Zhou, Y., Zhang, Y., Zhang, L. \& Hua, Z. DERD: data-free adversarial robustness distillation through self-adversarial teacher group. {\em Proceedings Of The 32nd ACM International Conference On Multimedia}. pp. 10055-10064 (2024)
\bibitem{bubeck2021universal}Bubeck, S. \& Sellke, M. A universal law of robustness via isoperimetry. {\em Advances In Neural Information Processing Systems}. \textbf{34} pp. 28811-28822 (2021)

\end{thebibliography}

\end{document}